\documentclass[journal]{IEEEtran}
\IEEEoverridecommandlockouts
\usepackage{graphicx}
\usepackage{amsmath}
\usepackage{multirow}
\usepackage{subfig}
\usepackage{caption}
\usepackage{bm}
\usepackage{amsbsy}
\usepackage{amsfonts}
\usepackage{amssymb}
\usepackage{amsmath}
\usepackage{color,soul}
\usepackage[ruled,vlined]{algorithm2e}
\usepackage{scrextend}

\usepackage{tikz}
\newcommand*\circled[1]{\tikz[baseline=(char.base)]{
		\node[shape=circle,draw,inner sep=1.0pt] (char) {#1};}}

\title{\LARGE \bf Swarm Control of Magnetically Actuated Millirobots}

\author{ Pouria Razzaghi, Ehab Al Khatib, and Yildirim Hurmuzlu
	\thanks{Pouria Razzaghi, Ehab Al Khatib, and Yildirim Hurmuzlu are with the Department of Mechanical Engineering, Southern
		Methodist University, Dallas, TX 75275, U.S.A. (email:
		hurmuzlu@lyle.smu.edu).}}

\begin{document}
	
		\maketitle
	\thispagestyle{empty}
	\pagestyle{empty}

\begin{abstract}

Small-size robots offer access to spaces that are inaccessible to larger ones. This type of access is crucial in applications such as drug delivery, environmental detection, and collection of small samples. However, there are some tasks that are not possible to perform using only one robot including assembly and manufacturing at small scales, manipulation of micro- and nano- objects, and robot-based structuring of small-scale materials. The solution to this problem is to use a group of robots as a swarm system. Thus, in this article, we focus on tasks that can be achieved using a group of small-scale robots. 

These robots are typically externally actuated due to their size limitation. Yet, one faces the challenge of controlling a group of robots using a single global input. In this study, we propose a control algorithm to position individual members of a swarm in predefined positions. A single control input applies to the system and moves all robots in the same direction. We also add another control modality by using different length robots. 

In our previous work \cite{Alkhatib2020}, we developed a small-scaled magnetically actuated millirobot. An electromagnetic coil system applied external force and steered the millirobots. This millirobot can move in various modes of motion such as pivot walking and tumbling. In this paper, we propose two new designs of these millirobots. In the first design, the magnets are placed at the center of body to reduce the magnetic attraction force between the millirobots. In the second design, the millirobots are of identical length with two extra legs acting as the pivot points. This way we vary pivot separation in design to take advantage of variable speed in pivot walking mode while keeping the speed constant in tumbling mode.

This paper presents a general algorithm for positional control of $n$ millirobots with different lengths to move them from given initial positions to final desired ones. This method is based on choosing a leader that is fully controllable. Then, the motions of a group of follower millirobots are regulated by following the leader and determining their appropriate pivot separations in order to implement the intended swarm motion. Simulations and hardware experiments validate these results. 
\end{abstract}

\section{Introduction}	

Global control of a population of robots is a challenging task that requires either on-board computation \cite{milutinovi2006modeling} or a broadcast signal \cite{shahrokhi2017steering}.
Swarm control of untethered small-scaled robots has recently become a popular research topic in the controls and robotics field. The size limitation of these robots makes on-board computation nearly impossible. Researchers have found ways to control groups of robots externally, such as applying a magnetic field \cite{zhang2018design}. Applying varied control inputs to individual tiny robots is also difficult. One solution is using a global control input that covers all robots. This means that a single actuation controls all robots. Moving and positioning a group of robots shows promising applications in fields such as biomedical engineering and biomechanics, particularly in drug delivery and tissue rehabilitation \cite{manshadi2018delivery,manshadi2019magnetic}. In this study, we focus on positioning a group of small-scale robots using a shared global control input. 


Applying the same control input to different robots results in the system being under-actuated. It means that we have a single input, but $n$ degrees of freedom for $n$ swarming robots. Other researchers have attempted to control this type of under-actuated system by adding extra constrains. These include placing obstacles in the workspace \cite{mahadev2017mapping,joshi2019motion}, providing  non-slip boundary contacts \cite{shahrokhi2017algorithms,shahrokhi2019exploiting}, changing the physical shape of the robot \cite{donald2008planar}, and applying an external artificial force field \cite{vose2012sliding}. When there is a large number of the robots, it can be difficult to detect their individual positions, however Shahrokhi \textit{et al.} showed that ``it is possible to sense global properties such as mean position and variance'' \cite{shahrokhi2017steering}. In this type of swarm control, a covariance ellipse was defined based on the most populated region of the workspace and the mean position is at the center \cite{shahrokhi2019planar}. Although they can place the mean position of the robots within the ellipse at the desired point, a number of robots outside the covariance ellipse can be missed or uncontrolled. Also, Dong and Sitti \cite{dong2020controlling} worked on a programmable and reconfigurable system as an external static magnetic field to control the formation of micro-robots. They experimentally showed that the swarm motion of these robots can manipulate the objects and navigate through complex environments.

Exciting articles that deal with swarm control and pattern formation algorithms are typically based on complex algorithms or introducing hard constrants in the workspace. Alternatively, our approach is based on simple algorithms and avoiding manipulation of the workspace. In this work, we control a group of robots under a unified control input. The robots respond differently to the same control signal due to their different physical structures. A different geometry (length) in the present work is utilized to add another degree of control modality to the system. The objective of swarm control is to move a group of robots from their initial to desired final positions, in which each robot is traceable. To achieve this objective, varying the length of the robots could be useful. Each set of desired final position and number of robots requires different sets of robot lengths.     

In our previous work \cite{Alkhatib2020}, we proposed a small-scaled robot (millirobot) that was actuated by an external magnetic field. A semi elliptical-shaped millirobot is built using 3D printing. A cylindrical permanent magnet is embedded in the center of the body. By changing the magnitude and the direction of the magnetic field vector, the millirobot can be actuated and moved in a specific direction. The motions are inspired by inertial actuation, which was developed in our lab \cite{zoghzoghy2015modeling,kashki2016pivot,razzaghi2019nonlinear}. Each millirobot can move in a variety of locomotion modes, such as pivot walking, tapping, and tumbling each with respective advantages and disadvantages (one can find more details in \cite{Alkhatib2020}). Among these modes, pivot walking is the fastest and most repeatable mode. Thus, we select the pivot walking as the primary mode in the present paper and the tumbling mode as the secondary one. When the global control input is applied to the millirobots, they will move parallel to each other but their velocities will be different in the pivot walking mode. The difference in their velocities would be proportional to their lengths. We exploit this feature to place an arbitrary number of millirobots, with pre-assigned lengths, at desired final locations. Also, the millirobots will move parallel and with the same velocities in the tumbling mode. This will give us an extra tool to move a swarm of millirobots.



\begin{figure}[h!]
	\includegraphics[width=1\columnwidth]{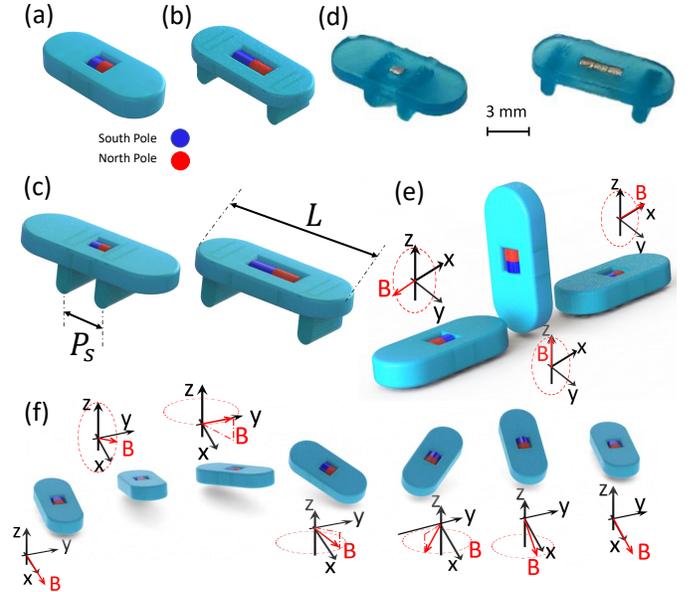}
	\centering
	\caption{\small{Millirobots and motion schemes of pivot walking and tumbling modes. \textbf{(a)} CAD design illustration of a millirobot without legs (primary design). The permanent magnet is embedded at the center of the body. \textbf{(b)} CAD design illustration of a millirobot with two legs (secondary design). The red and blue colors represent the north and south poles of a magnet. \textbf{(c)} Illustrations of two legged millirobots with different pivot separations. $L$ and $P_s$ denote the length and pivot separation, respectively. \textbf{(d)} Actual 3D printed two millirobots with different pivot separations. \textbf{(e)} The sequences of tumbling motion; one tumbling step is achieved by rotating the magnetic vector about the y-axis by $180^{\circ}$. \textbf{(f)} The sequences of pivot walking motion; the scheme shows a complete locomotion step. It is achieved by lifting one end and forming a pivot point at the other end by rotating the magnetic vector around the y-axis;  rotate the magnetic vector about the z-axis to rotate the millirobot about the formed pivot as sweep angle; the process is then repeated in the opposite direction. }} 
	\label{Fig2}
\end{figure}

In this paper, we propose two different designs of millirobots. A stadium shape with a cylindrical permanent magnet embedded at the center of the body is the primary design (see Fig.~\ref{Fig2}(a)). We should note that placing the magnets at the center of the body reduces the attraction forces between the magnets that appeared in our previous millirobots \cite{Alkhatib2020}.  The new millirobots are printed in four different lengths as 3, 5, 7, and 9 mm. In this design, the velocities of steering the millirobots are proportional to their lengths in both pivot walking and tumbling modes of motions. In order to differentiate pivot walking and tumbling motions, we change the design by adding two legs (see Fig.~\ref{Fig2}(b)). In the secondary design, the lengths of the millirobots are fixed at 10 mm, but they have different pivot separations between two legs ($P_s$) as 3, 5, 7, and 9 mm. Thus, they can move in different velocities in the pivot walking mode, but the same velocity in the tumbling mode. This difference gives us two fundamental flexibilities in the proposed swarm pattern motion. First, we can generate a specific and desired formation employing the swarm algorithm in the pivot walking mode; then, the formation can be moved to any desired location using the tumbling mode without any changes in the final shape. The illustration of the pivot walking and tumbling motions are shown Figs.~\ref{Fig2}(d) and (e). Also, the directions of the magnetic field required to conduct each motion are drawn.  
              
\normalsize

\begin{figure}[h!]
	\includegraphics[width=1\columnwidth]{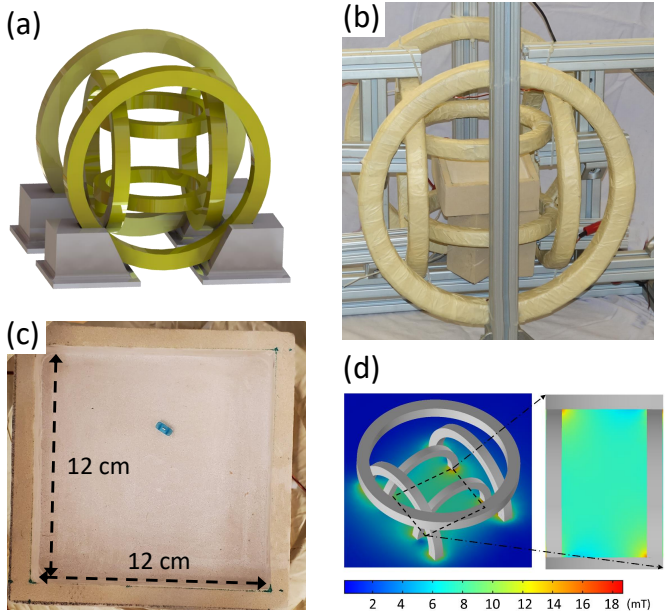}
	\centering
	\caption{\small{ A nested Helmholtz electromagnetic coil system.
			\textbf{(a)} Isometric view of CAD design. \textbf{(b)} Isometric view of actual system. \textbf{(c)} Top view of workspace. \textbf{(d)} Simulation result of magnetic field at the center of the workspace. }} 
	\label{Fig3}
\end{figure}

A nested electromagnetic Helmholtz coil is designed and constructed to actuate the presented millirobots. This system is configured based on the optimal design presented in \cite{abbott2015parametric}. The large-scale coil system produces an uniform static magnetic field, which can rotate in 3D dimension. The outer diameters of coils are 39, 30.5, and 22.5 cm in x, y, and z directions respectively. The separation distances between coil pairs are 24, 19, and 11 cm. The system has a $12\, \text{cm}\times 12\, \text{cm}$ work space at the center of the configuration (see Fig.~\ref{Fig3}(c)). The coils are fabricated using insulated 12 gauge circular copper wire. Figures~\ref{Fig3}(a) and (b) show the isometric views of the CAD drawing and the actual coil system. The maximum current applied to the system is eight amps and the system can generate a continuous magnetic fields above 10 millitesla (mT). We simulate the magnetic field profile at the center of the configuration using \textit{Comsol} software as shown in Fig~\ref{Fig3}(d).


\normalsize

\section{Mathematical Model}

In the locomotion of the system, we assume that two ends of the robot's body (first design) and two legs (second design) are acting as the pivot points. A stationary electromagnet system produces a uniform rotating magnetic field in three dimensions. This rotating magnetic field generates torques on the magnets embedded into the millirobot. This aligns the long axis of the body with the applied magnetic field vector \cite{Alkhatib2020}. Thus, we are able to translate the center of mass of the body and perform rotations about the in-plane and out-of-plane angles. 

\subsection{Pivot Walking}
Pivot walking is achieved by successively alternating the direction of the magnetic field vector in the positive and negative $z$-directions and rotating around $z$-axis as shown in Fig.~\ref{Fig2}(c). When the magnetic field vector is oriented in the positive $z$-direction, the induced magnetic torque presses one end down while the other end is lifted up. Subsequently, while having a pivot formed at the pressed end, a positive rotation about $z$-axis is applied. This causes the millirobot to rotate forward by a sweep angle of $\theta_{i}$ in the $x-y$ plane in its i$^{th}$ step. In the next step, the orientation of the magnetic field in $z$-direction is reversed, and the pivot moves to the other end. A negative rotation about $z$-axis is applied to rotate the millirobot by $\theta_{i+1}$ about the new pivot point. We consider this process as a complete step. Repeating this process, locomotion along a desired path is generated. Also, a single tumbling motion step is achieved by a rotation of the magnetic field vector about $x$-axis by a 180 $\deg$ as shown in Fig.~\ref{Fig2}(e).

\begin{figure}[h!]
	\includegraphics[width=1\columnwidth]{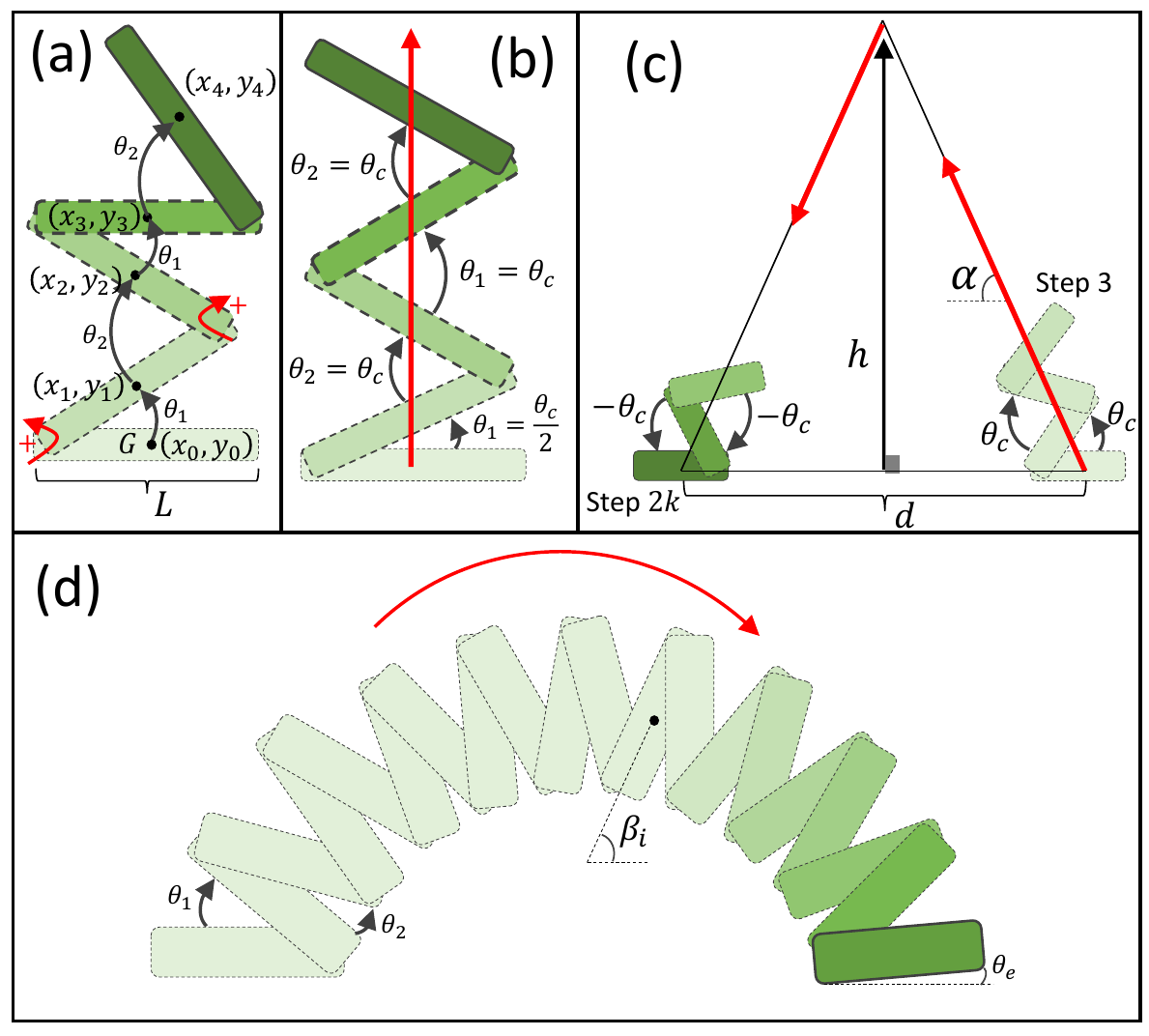}
	\centering
	\caption{\small{Millirobot walking tools \textbf{(a)} The schematic of the pivot walking with coordinates of the center of body and the positive directions of sweep angles. \textbf{(b)} Straight line motion. \textbf{(c)} Triangular trajectory path. \textbf{(d)} Circular path configuration.}} 
	\label{Figtools}
\end{figure}

Here, we calculate the coordinates of the center of mass $(x_k,y_k)$, with $k$ being the number of steps. The kinematic modeling of pivot walking depicted in Fig.~\ref{Figtools}(a) can be expressed as follows:
 
\footnotesize{ \begin{flalign}
 x^n_k=x^n_0+&\frac{L_n}{2}\sum_{i=1}^{k}\left( (-1)^i \cos \left[ (-1)^i \left \lfloor \frac{i}{2} \right \rfloor \theta_1 + (-1)^{i-1} \left \lfloor \frac{i-1}{2} \right \rfloor \theta_2  \right] \right.  \nonumber \\
&+ \left.(-1)^{i-1} \cos \left[ (-1)^{i-1} \left \lfloor \frac{i+1}{2} \right \rfloor \theta_1 + (-1)^{i} \left \lfloor \frac{i}{2} \right \rfloor \theta_2  \right]   \right )  \label{Eq3}\\
y^n_k=y^n_0+&\frac{L_n}{2}\sum_{i=1}^{k}\left( (-1)^i \sin \left[ (-1)^i \left \lfloor \frac{i}{2} \right \rfloor \theta_1 + (-1)^{i-1} \left \lfloor \frac{i-1}{2} \right \rfloor \theta_2  \right] \right.  \nonumber \\
& +\left.(-1)^{i-1} \sin \left[ (-1)^{i-1} \left \lfloor \frac{i+1}{2} \right \rfloor \theta_1 + (-1)^{i} \left \lfloor \frac{i}{2} \right \rfloor \theta_2  \right]   \right ) \label{Eq4}
\end{flalign} }\normalsize
where ($x_0,y_0$) are the coordinates of the initial position of the millirobot, $n$ denotes the number of millirobot, $L$ is the length, and $(\theta_1, \theta_2)$ are the sweep angles around two pivot points, respectively. Also, the $\left \lfloor . \right \rfloor$ denotes the floor function, which is the function that takes as input a real number and gives as output the greatest integer less than or equal to the input. 

\subsection{Basic Motion Paths}
One can obtain different motion paths by choosing different combinations of sweep angles. Three basic motion paths explored here as tools for swarm control; straight, triangular, and circular paths. Figure~\ref{Figtools} shows the schematic representations of these paths. The straight-line motion shown in Fig.~\ref{Figtools}(b) is generated by choosing same sweep angles for each pivot step $(\theta_1=\theta_2= \theta_c)$, however, the first sweep angle has to be the half of others $(\theta^1_1 \, \text{or}\, \theta^2_1= \theta_c/2)$. The distance covered by millirobot, in this case, is proportional to the length of the millirobot. 

A triangular trajectory is achieved by selecting equal sweep angles for the first k$^{th}$ steps ($\theta^1_{1:k}=\theta^2_{1:k}= \theta_c$) and the sweep angles are switched to negative $\theta_c$. The two sets of steps are considered to be a complete step. If the millirobot starts from a line, after $ 2k $ steps (end by a complete step), it goes back to the same line (see Fig.~\ref{Figtools}(c)). The trajectory is an isosceles triangle, and equal base angles can be expressed in terms of the sweep angle $(\alpha= \frac{\pi-\theta_c}{2})$. The base $(d)$ and height $(h)$ of this triangle are related to the length of the millirobot, the sweep angle, and number of steps as follows:
\begin{flalign}
& h=\frac{d}{2} \cot \left(\frac{\theta_c}{2}\right) \\
& y_k= h
\end{flalign} 
 
In order to follow a circular path, two sweep angles must be different $(\theta_1\neq\theta_2)$. The radii of the generated circle is related to the sweep angles and length of millirobot. The equation of this trajectory can be found as:
\begin{flalign}
&  x_i^2+y_i^2=r^2_c \\& y_k=y_0
\end{flalign}
where $r_c$ denotes the radius of circular path (see Fig.~\ref{Figtools}(d)). If the millirobot starts from a line, after $2k-1$ steps, it is not guaranteed to return to the same line. Thus, the sweep angle of the last step should be performed with a different sweep angle. 

Consider a circular path with $ 2k $ steps. The millirobot moves $ k $ steps with sweep angle $\theta_1$ on the first pivot point and $ k-1 $ steps with sweep angle $\theta_2$ on the second pivot point. The angle between the long axis of the body and the positive direction of $ x $-axis in each step ($\beta_i$) and the extra sweep angle ($\theta_e$) to complete the round (last step or $2k^{th}$ step), can be calculated as:
\begin{flalign}
&\theta_d=\left \lfloor \frac{i+1}{2} \right \rfloor \theta_1-\left \lfloor \frac{i}{2} \right \rfloor \theta_2  \\
& \beta_i= \left\{\begin{matrix}
\theta_d & \theta_d\leq 90^{\circ}\\  
180^{\circ}- \theta_d& \theta_d > 90^{\circ}
\end{matrix}\right. \\
& \theta_e = 180^{\circ}-\left(k\, \theta_1-(k-1)\theta_2\right)
\end{flalign}  
  
We utilize these trajectories as tools to conduct swarm positioning control.

\subsection{Swarm Motion Using Basic Motion Paths} \label{Swarm2}
\begin{figure}[h!]
	\includegraphics[width=1\columnwidth]{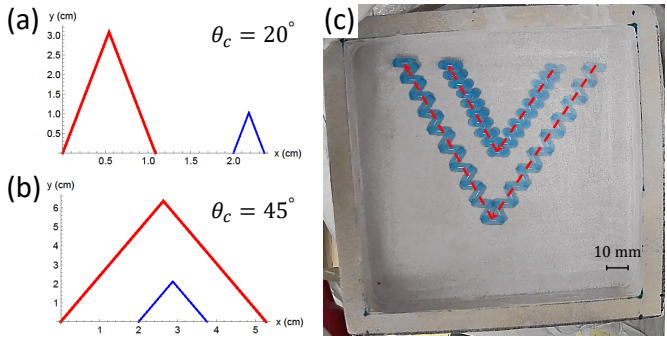}
	\centering
	\caption{\small{Changing the final distance between two millirobots. The initial distance between two millirobots is $\Delta \,r=2 \,\text{cm}$ and the lengths of the robots are 1.5 cm for red trajectory and 0.5 cm for blue one. In both paths, the millirobots change the direction after 8 steps. \textbf{(a)} The sweep angle is $\theta_c=20^{\circ} $ and the distance after at the end is approximately 1.25 cm. \textbf{(b)} The sweep angle is $\theta_c=45^{\circ} $ and the distance at the end is approximately 1.4 cm, but the order of the millirobots is changed.  \textbf{(c)} Experimental results of the position altering of two millirobots. The pivot separations are 5 and 9 mm. }} 
	\label{Fig31}
\end{figure}

Here, we consider two millirobots with different pivot separations. Initially they are placed on a straight line with a separation of $\Delta \,r$ between them. By using triangular path planning, one can change the distance between the millirobots and reverse their initial order on the original line (see Fig.~\ref{Fig31}).

\begin{figure}[h!]
	\includegraphics[width=1\columnwidth]{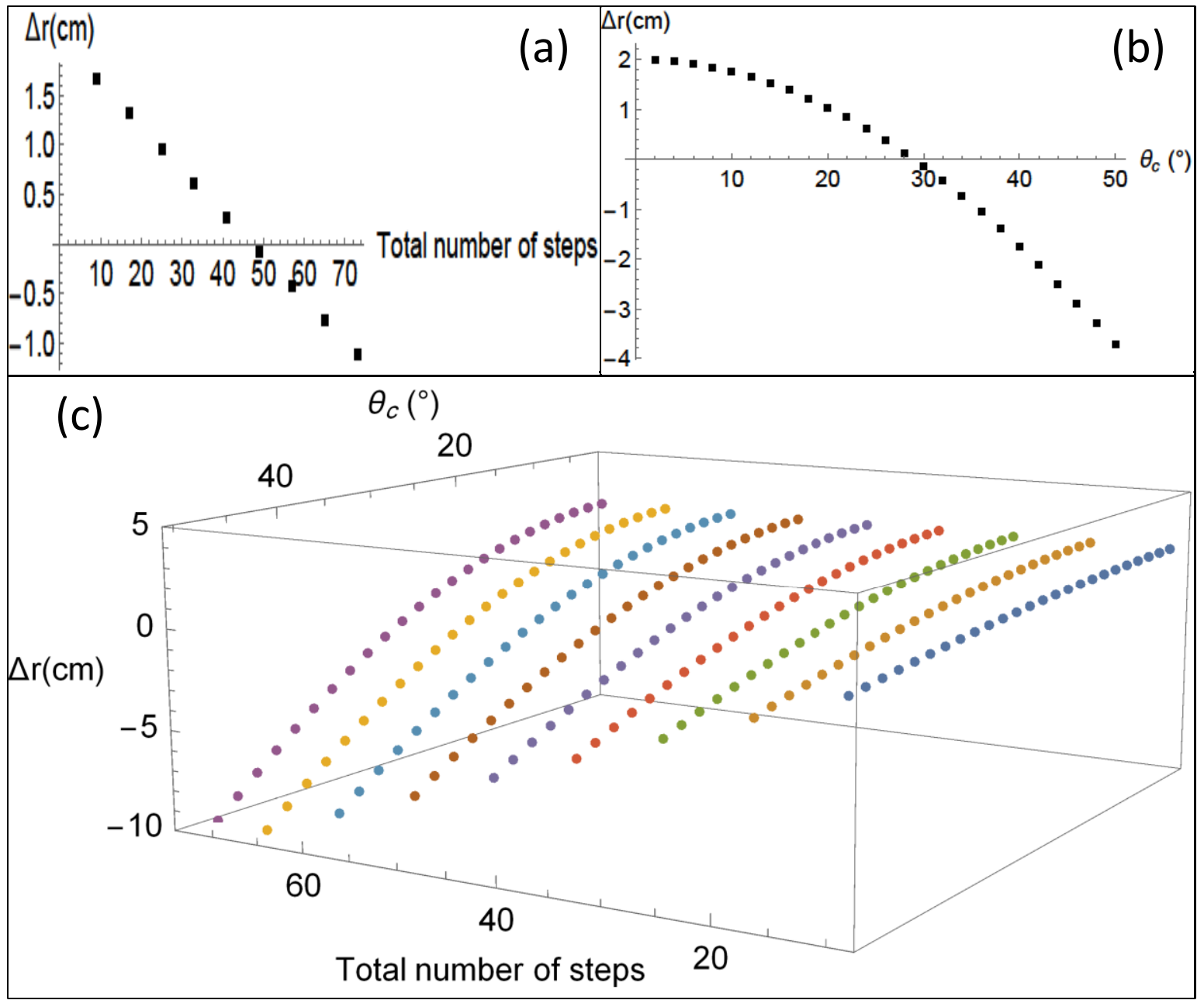}
	\centering
	\caption{\small{Variation of the final distance between two millirobots with lengths of 2 and 1 cm for different sweep angles and number of steps,  when the initial distance is 2 cm. \textbf{(a)} The sweep angle is fixed at $\theta_c=24^{\circ} $ and the number of steps is varied. \textbf{(b)} The sweep angle is varied when the number of steps sets at 33. \textbf{(c)} Varying both sweep angle and number of steps.}} 
	\label{Fig4}
\end{figure}

To express the position alteration of two millirobots more accurately, we conduct a parametric analysis of the effects of sweep angle and number of steps on the final positions (see Fig.~\ref{Fig4}). Figure~\ref{Fig4}(a) depicts the effect of the total number of steps on the final distance between two millirobots when a constant sweep angle $\theta_c=24^{\circ} $ is used. A negative value for distance means the order of two millirobots is preserved. Also, Fig.~\ref{Fig4}(b) shows the alteration in the relative position of millirobots in terms of changing sweep angles in a constant number of steps $2 k+1 = 33$. The direction of the path is altered at $k=12$. In Fig.~\ref{Fig4}(c), one can see the difference in distance between two millirobots at the end of the triangular path motion as a function of the sweep angle and number of steps (see \textbf{SP}~1).  

Based on the results shown in Fig.~\ref{Fig4}, we can claim that with a specific combination of sweep angle and number of steps, any two millirobots with different pivot separation can walk to final positions with their relative distance set to an arbitrary desired value. This claim will be proved in Section~\ref{Controllability}. Here, we present a formal mathematical formulation for this type of action; consider two millirobots with different lengths $L_1$ and $L_2$ starting on a line with a relative distance of $\Delta\,r$. Whereas, $\Delta\,p$ denotes the final desired value of the relative distance between two millirobots. From Eq.~(\ref{Eq3}), the base of the triangle can be found as:
\begin{equation}
d= |x_k^n-x_0^n|=f_x(L_n, \theta_c,k)
\end{equation}  
where $f_x(.)$ is a function of three parameters ($L_n,\theta_c,k$), which can be extracted from the right hand side of Eq.~(\ref{Eq3}) and the combination set can be expressed as follows:
\begin{equation}
S=\{(\underset{i=1,2}{L_i},\theta_c,k) \: \:| \:\: \underset{i=1,2}{d_i= f_x(L_i, \theta_c,k)} , d_2=\Delta\,r+d_1\pm \Delta\,p\}
\end{equation}
where $d_1$ and $d_2$ are the bases of triangular paths for each millirobot. One can use this motion path to change the order of any number of millirobots initially placed on a straight line. Subsequently, we conduct an experiment to show this ability only for three millirobots with different pivot separations due to the restriction imposed by the size of the workspace (see Fig.~\ref{ChangeOrder3} and \textbf{SP}~1). We should note that in the following figures of the experimental results, we just show a select number of steps in the overlay pictures to highlight the overall path of the swarm motions without overcrowding the figures. One can see the experiments in the videos provided in the supplementary materials.   

\begin{figure}[h!]
	\includegraphics[width=0.8\columnwidth]{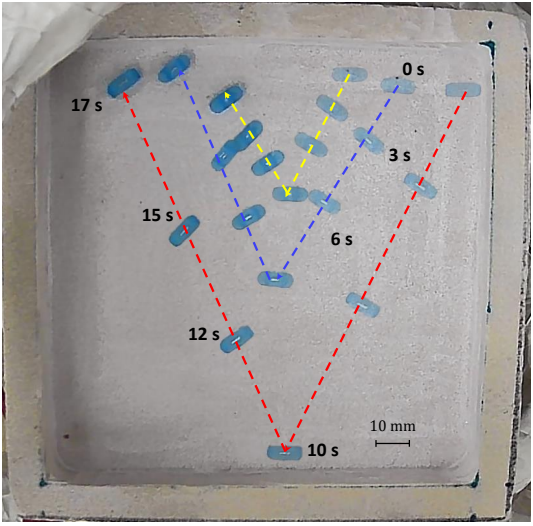}
	\centering
	\caption{\small{Sequences of changing order of three millirobots with 3, 5, and 9 mm in pivot separations. This maneuver is performed in 30 steps. The millirobots are approximately moving two steps per second. }} 
	\label{ChangeOrder3}
\end{figure}
    
\begin{figure}[h!]
	\includegraphics[width=1\columnwidth]{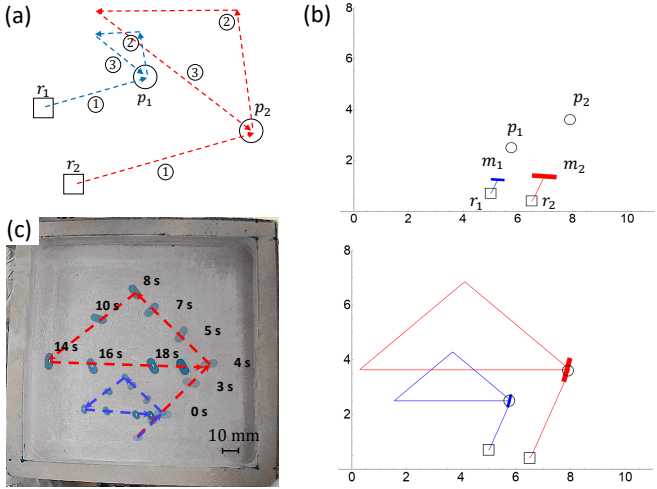}
	\centering
	\caption{\small{Swarm pattern motion of two millirobots with 3 and 9 mm lengths. \textbf{(a)} Illustration of swarm  motion. Initial positions are marked by squares and final ones by circles. Dashed lines show the paths of the centers of millirobots in three runs. \textbf{(b)} Simulation result of walking two millirobots. \textbf{(c)} The experimental result of the swarm motion of two millirobots.}} 
	\label{Fig5}
\end{figure}

We present a swarm motion of two millirobots using the basic paths. This swarm motion shows the capability of the walking tools. We assume that the initial and final positions of two millirobots are $(r_1,r_2)$ and $(p_1,p_2)$ respectively. Without loss of generality, we consider that the length of millirobot 2 $(m_2)$ is greater than millirobot 1 $(m_1)$ ($L_1<L_2$). Thus, $m_2$ moves faster and undergoes longer triangular path. This motion consists of three walking runs including a line with a slope, triangular path, and straight line motions to generate a swarm pattern motion. Figure~\ref{Fig5} shows an illustration of simulation and experimental results of the swarm motion of two millirobots with different lengths (see \textbf{SP}~2).  
 
\section{Controllability Analysis} \label{Controllability}

The kinematic equations of motion of the $\text{i}^{th}$ robot can be represented using a unicycle model as follows:
\begin{align} 
\dot{x}_i &=  u_r \cos {(\theta + \frac{\pi}{2})} +  u_{p_i} \cos {\theta} \label{xequation}\\
\dot{y}_i &=  u_r \sin {(\theta + \frac{\pi}{2})} +  u_{p_i} \sin {\theta} \label{yequation}
\end{align}
where $u_t= K_t L_r $ is the tumbling speed and $u_{p_i}=K_p L_{p_i}$ is the pivot walking speed. The terms $K_t$ and $K_p$ are speed constants in tumbling and pivot walking modes. The $L_r$ and $L_{p_i}$ denote the length of robot and pivot separation respectively. One can write Eqs.~(\ref{xequation}) and (\ref{yequation}) in matrix form as:
\begin{flalign}
&\dot{\bm{x}}_i = \bm{A} \bm{x}_i + \bm{B}_i \bm{u} \overset{\text{where}}{\rightarrow} \bm{x}_i = \begin{bmatrix}
x_i  \\ 
y_i
\end{bmatrix} \\
& \bm{A} = \begin{bmatrix}
0 &0 \\ 
0 & 0
\end{bmatrix} ;\, \,\, \bm{B}_i =  \begin{bmatrix}
0 & \nu_i & -1 & 0\\ 
\nu_i & 0 & 0 & 1
\end{bmatrix} \nonumber \\
& \text{and} \, \bm{u}= 	\begin{bmatrix}
K_p L_r \sin \theta\\ 
K_p L_r \cos \theta\\ 
K_t L_r \sin \theta \\ 
K_t L_r \cos \theta
\end{bmatrix} ;\text{where} \,\, \nu_i=L_{p_i}/L_r\nonumber
\end{flalign}

This model can be generalized to describe the swarm system as follows:
\begin{flalign}
& \dot{\bar{\mathbf{x}}} = \mathbf{A} \bar{\mathbf{x}} + \mathbf{B} \bar{\mathbf{u}} \\
& \mathbf{x} = \begin{bmatrix}
x_1 & y_1& \cdots & x_n & y_n
\end{bmatrix}^T_{2n \times 1} \nonumber
\end{flalign}
where $\mathbf{A}$ is $2n \times 2n$ zero matrix, and $\mathbf{B}$ is a $2n \times 4$ matrix represented as:
\begin{flalign}
 \mathbf{B} = \begin{bmatrix}
0 & \nu_1 & -1 & 0\\ 
\nu_1 & 0 & 0 &1 \\ 
\vdots& \vdots & \vdots  &\vdots \\ 
0 & \nu_n & -1 & 0\\ 
\nu_n & 0 & 0 &1 \\ 
\end{bmatrix}_{2n \times 4} 
\end{flalign}

The controllability matrix $\mathbb{C}$ of the swarm system can formulated as:
\begin{equation}
\mathbb{C} = \begin{bmatrix}
B, AB, A^2B, \cdots, A^{2n-1}B
\end{bmatrix}
\end{equation}

For a system with only one robot, the rank of $\mathbb{C}$ is two, which means all degrees of freedom (DOF) are controllable. For a swarm system with two robots with different lengths ($\nu_1 \neq \nu_2$), the rank of matrix $\mathbb{C}$ is four, which means the system is controllable. This capability of swarm system for two robots is numerically and experimentally shown in Section~\ref{Swarm2} and Fig.~\ref{Fig5}. We showed that the robots are steered into their corresponding desired final positions from arbitrary initial positions.

For swarm systems with more than two robots, the rank of $\mathbb{C}$ will be still four, which means only four DOFs are controllable. Hence, the motions of other robots are related to the motion of the fully controllable ones. This is exactly the basis of our swarm algorithm. 

In a $n$-millirobot swarm system with different lengths, there are $2n$ DOFs, which the controllability analysis shows that only four of them are controllable. One can select any four DOFs and control them separately. In this study, we choose only one controllable millirobot and propose a swarm algorithm based on its controlled DOFs. 

It is worth to mention that the controllability analysis results are directly tied to the fact the external input translate into applying the same rigid body transformations to move the millirobots in tumbling and pivot walking modes. The pivot walking is scaled by the constant $\nu_i$ for each robot. This parameter linearly depends on the pivot separation or the length of each millirobot.

\section{Swarm Position Control of $n$ Millirobots}

The proposed swarm control methodology often requires a priori determination of the lengths of the millirobots required to perform a specific placement task. In this section, we present an algorithm that yields the required robot lengths, sweep angles, and number of steps to move millirobots from their initial positions to desired final destinations.

Algorithm~1 is used to position $n$ millirobots from initial positions to final desired ones. This positional control of $n$ millirobots can be carried out by calculating the lengths of millirobots based on their initial positions and desired final destinations. The algorithm presents the process of finding different sets of $(L_n, \theta_n,k_n)$ to perform this task. 

\begin{algorithm} [h!]
	\SetAlgoLined
	\textbf{Input parameters:} The coordinates of initial positions $(x_0^i,y_0^i\,;i=1:n)$ and final destinations $(x_f^i,y_f^i \,;i=1:n)$.  \\
	\textbf{Calculation:} \\
	1: Set length of $m_1$ as $L_1$ and move it to its final position $p_1$ \\
	2: From Eqs.~(\ref{Eq3}) and (\ref{Eq4}), solve for the first set: \\ \small{$S_1=\{(\theta_{c1},k_1) \, | \, \text{Straight motion} \, , \, (x_0^{1},y_0^{1}) \rightarrow (x_f^{1},y_f^{1})\}$ } \normalsize  \\
	3: Use set $S_1$, and calculate the lengths of other millirobots: \\
	\small{$S_i=\{(L_i) \, | \, \theta_1=\theta_2=\pm \theta_{c1} \, , \, (x_0^{i},y_0^{i}) \rightarrow (x_f^{i},y_f^{i}) , i=2:n\}$ } \normalsize  \\ 
	6: Swarm set: $S_{swarm}= \bigcup_{i=1}^{n} S_i $
	\caption{Swarm control-$n$ millirobots}
	\label{Alg2}
\end{algorithm}
 
In order to position $n$ millirobots at their corresponding final destinations, one can use Algorithm~\ref{Alg2}. Let's consider the first millirobot $(m_1)$ as it moves to its final position with straight motion based on Eqs.~(\ref{Eq3}) and (\ref{Eq4}), by calculating the set $S_1=\{(\theta_{c1},k_1) \, | \, \text{straight motion} \, , \, (x_0^{1},y_0^{1}) \rightarrow (x_f^{1},y_f^{1})\}$. Using this set and applying it to Eqs.~(\ref{Eq3}) and (\ref{Eq4}) for other millirobots, the coordinates of midpoint position of each millirobot $(x_1^i,y_1^i\,; i=2:n)$ can be expressed as a function of their lengths $(L_i \,; i=2:n)$, sweep angle, and number of steps. The relative distance between the initial and final position of a midpoint of each millirobot is also a function of its length, sweep angle, and number of steps. By solving the resulting equations, one can get the sets of millirobot lengths, sweep angles, and the number of steps off-line. Then, the millirobots can be placed at their initial positions and the swarm control can be conducted by applying the solution sets.

\section{Results}
 
\subsection{Swarm Pattern Motions}

\begin{figure}[h!]
	\includegraphics[width=0.75\columnwidth]{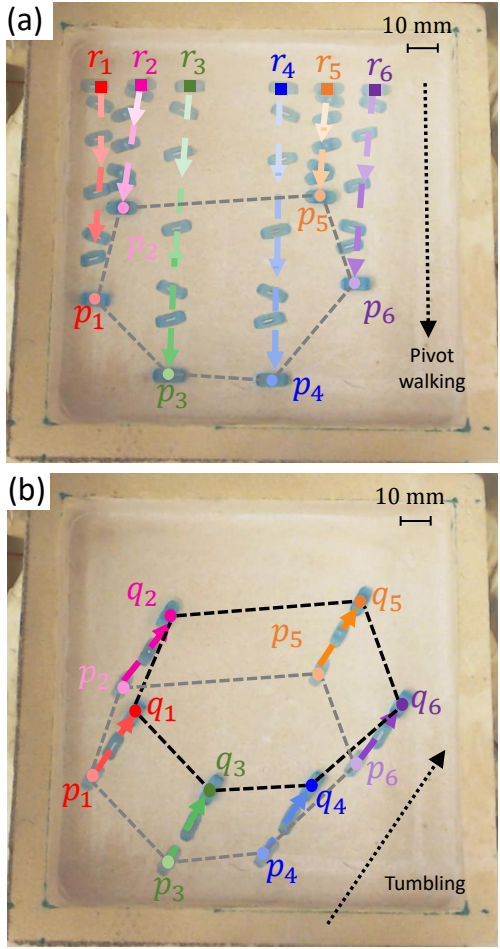}
	\centering
	\caption{\small{Swarm motions of six millirobots to generate a hexagon pattern. Each dashed color-line represents the path of the midpoint of each millirobot. The initial and final positions are shown by squares and circles symbols. \textbf{(a)} Six millirobots start from their initial positions ($r_i; \:\: i=1:6$) and move to points $p_i; \:\: i=1:6$ in pivot walking mode to generate a hexagon pattern. \textbf{(b)} The millirobots are steered to their final positions ($q_i; \:\: i=1:6$) in tumbling mode while the hexagon pattern is preserved. }} 
	\label{patterns}
\end{figure}

In this section, we experimentally demonstrate the swarm position control of the millirobots to generate different geometrical shapes including triangle, square, pentagon, and hexagon patterns (see \textbf{SP}~3). The edges of each shape are considered as the final desired positions of each millirobot. In these experiments, we use the secondary design of the millirobots, in which the pivot separations are fixed at 3, 5, 7, and 9 mm. By using the reverse solutions of Algorithm~1, one can find the desired initial positions to perform a swarm motion. In addition, we use the tumbling motion to move the final shape of patterns. This capability is the strength of the presented swarm motion. Any desired pattern can be generated through Algorithm~1 and then the final pattern can be placed anywhere by using the tumbling mode motions. Figure~\ref{patterns} depicts the experimental result of the swarm position motions to generate a Hexagon pattern. This swarm motion is conducted by six millirobots with three different lengths. We use two millirobots of each length, which are 3, 7, and 9 mm. The millirobots start at points $r_i; \:\: i=1:6$. They move to points $p_i; \:\: i=1:6$ in pivot walking mode to generate the desired hexagon pattern. Then, they are steered in tumbling mode to reach their corresponding final points $(q_i; \:\: i=1:6)$. Also, it should be noted that in the figure we only show the experimental results of the hexagon pattern to reduce the complexity (the videos are included in the supplementary materials for the information of the readers).

\subsection{Swarm application}
The objective of swarm control of a group of robots is to perform a task, which is not possible to perform with a single robot. Here, we numerically and experimentally show extra practical applications of the swarm motions of these millirobots. We should note that we assume that there are no collisions among the millirobots during their motions or any interaction between them.   
\subsubsection{Expansion maneuver}
\begin{figure}[h!]
   \includegraphics[width=0.75\columnwidth]{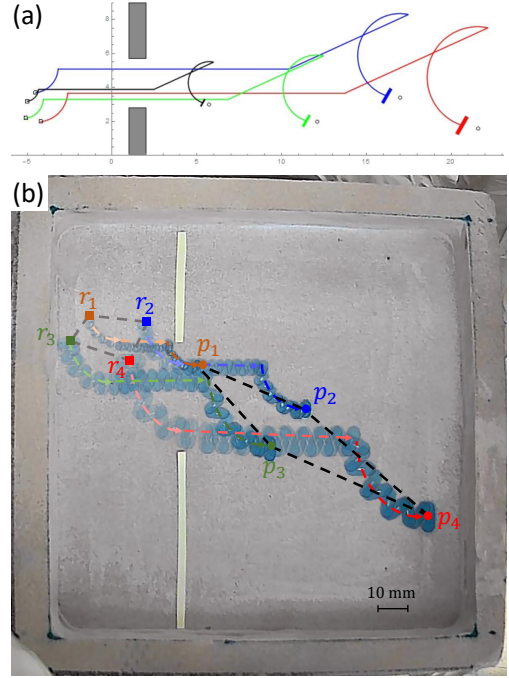}
   \centering
   \caption{\small{Sequences of the horizontal expansion maneuver of four millirobots with different lengths as 3, 5, 7, and 9 mm. The compacted initial and the final expanded formations are indicated by gray and black dashed polygons respectively. Each dashed line represents the path of the midpoint of each millirobot. }} 
   \label{Fig8}
\end{figure}

A group of four primary-design millirobots conducts a maneuver that expands from a contracted initial formation to an expanded one. We call this maneuver ``Expansion''. The simulation of the expansion maneuver is conducted as follows; the group starts from a relatively compact of initial positions. Then, a circular motion is performed to place the robots in front of a narrow opening while fixing their relative distances at specific values. This formation makes it possible for the group to go through the channel by using straight-line motion. After passing the channel, depending on whether we require horizontal or vertical expansion, the subsequent scenarios are different. For instance, after formation undergoes the walking sequences to pass through the channel, they are steered on an inclined straight line to adjust their positions on subsequent circular paths. In the end, a circular motion is carried out to bring all millirobots to their final positions. (see Fig.~\ref{Fig8} (a)). We experimentally demonstrate the expansion maneuver conducted by four millirobots. They are placed in specific compact initial positions to minimize the attraction forces between the magnets. Figure~\ref{Fig8}(b) shows a selected sequence of frames of experimental results of the expansion maneuver. The millirobots start from their initial positions ($r_i; \:\: i=1:4$). They move in a circular path to a location in front of the opening; then, walk in a straight line formation to go through the channel. In the end, circular and line motions are performed to expand the formation and reach final destinations ($p_i; \:\: i=1:4$) (see \textbf{SP}~4).

\subsubsection{Contraction maneuver} 
A group of millirobots is placed on the respective expanded formation of initial positions. Then, they undergo a reverse sequence of motions explained in the expansion maneuver. They are steered to the intermediate compact positions in front of a narrow channel and walked through it. In the end, a maneuver is performed in order to place the millirobots in respective desired final positions. Alternatively, the contraction maneuver can be named as the reverse maneuver of expansion motions.   
    
\subsubsection{Reverse maneuver} 

\begin{figure}[h!]
 	\includegraphics[width=0.75\columnwidth]{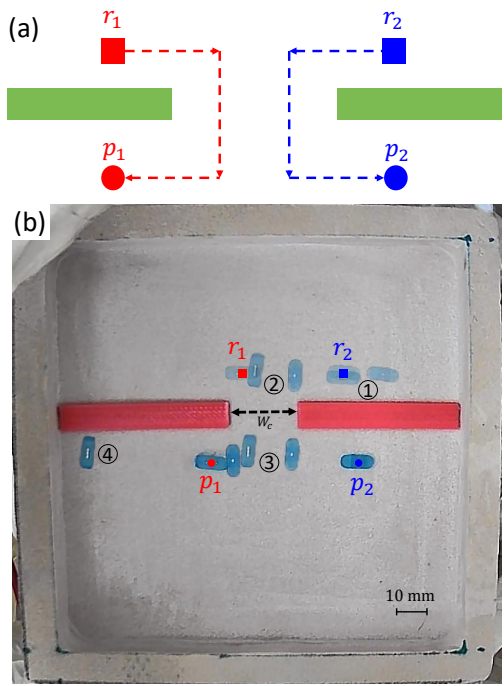}
 	\centering
 	\caption{\small{Sequences of the reverse maneuver of two millirobots with different pivot separations as 3 and 9 mm. (a) The scheme of the reverse maneuver. (b) The experimental result. }} 
 	\label{Fig10}
 \end{figure}

Under the same control input, all millirobots move in the same direction. A challenging task is to steer the millirobots in opposite direction by applying the same magnetic field. We propose a walking plan to move two millirobots from a set of initial positions to the desired set of final destinations, which is required an opposite direction motion. We name this maneuver ``Reverse''. Figure~\ref{Fig10}(a) shows an illustration of this maneuver. This walking plan consists of five sequences; including two pivot walking and three tumbling mode motions. The millirobots are placed on the top side of the two obstacles. They should pass a narrow channel with a width of $W_c$ and be positioned at the other side of the obstacles. 

We conduct an experiment to show the reverse maneuver as follows (see Fig.~\ref{Fig10}(b)); two millirobots with different pivot separations (3 and 9 mm) are placed at their corresponding initial positions ($r_1,r_2$). Here, the millirobot placed at $r_1$ has bigger pivot separation than the other one. First, they move to the point $\circled{1}$ in pivot walking mode to reduce the relative distance between them until it reaches less than $W_c$. Then, they tumble back to the point $\circled{2}$, as the relative distance between the millirobots remains constant. They tumble through the channel and pass it to the point $\circled{3}$. Then, they are steered to the point $\circled{4}$ in the pivot walking mode to increase the relative distance between them. At the end, they tumble to their final destinations ($p_1,p_2$) (see \textbf{SP}~5).
 
We should note that the experiments are conducted on a dry surface. Thus, we observe slippages at the pivot points in different situations. As we have pre-defined control inputs acting on the system, we can not overcome the effects of the slippages on the final outputs. This causes the experimental misplaced positions of millirobots from their simulated and desired final destinations. There are not any trends or similarities for the misplacement and are due to un-modeled friction on the surface. The solution for this problem would be to add a closed-loop controller to correct the motion of each millirobot, which will be proposed in our next study.  
 \section{Conclusion}
 
This study proposed a pre-computational technique for swarm position control of a group of small-scale robots using uniform input. The paper presented an algorithm for the positioning $n$-robots actuated by a uniform magnetic force field. The unique millirobots introduced in our previous study \cite{Alkhatib2020} modified in design. We placed magnets at the center of the body to reduce the magnetic attraction forces. Also, we added two legs acting as pivot points. In new design, by varying the pivot separation and keeping identical lengths, the millirobots can move in different velocities in pivot walking mode and constant velocity in tumbling mode. To obtain different positional outcomes out of steering millirobots under the same control input, we used millirobots with different lengths as well as variable pivot separation in pivot walking mode. 

First, we presented two modified designs of millirobots and listed their advantages. Then, we demonstrated different walking tools, which were utilized for the swarm motions. We analyzed the controllability of the swarm system and showed that up to two millirobots are fully controllable. We used only one controlled millirobot as the leader and developed an algorithm to place $n$ millirobots that follow the leader while moving from arbitrary initial positions to other arbitrary final positions. Accordingly, the required lengths of the follower millirobots were calculated based on the coordinates of the initial and final positions and a pre-computational path planning to perform the swarm motions. We verified the proposed algorithm for swarm positioning the millirobots through simulation and experiments. Also, we conducted different experiments to show the capability of our millirobots to perform a group task.

There were position errors in the experimental results due to slippage effects. For the next step, we are working on a closed-loop control strategy to conduct more precise experiments. 
 
\section*{Acknowledgments}
We would like to thank Mr. Necdat Yildimier for his help in fabricating the magnetic coils and its holding frame and also, Mr. Mohammad Karim Dehghan Manshadi for his guidance in preparing the \textit{Comsol} simulations.  
 
\section*{Supplementary materials} 
One can find the following videos as supplementary materials.
\begin{addmargin}[2em]{1em}
	\begin{itemize}
		\item[\textbf{SP} 1 :] Changing order of two and three millirobots.
		\item[\textbf{SP} 2 :] Swarm motion of two millirobots
		\item[\textbf{SP} 3 :] Swarm motion to generate different geometrical shapes.
		\item[\textbf{SP} 4 :] Expansion maneuver.
		\item[\textbf{SP} 5 :] Reverse maneuver.
	\end{itemize}
\end{addmargin}

\bibliographystyle{IEEEtran}
\bibliography{paper}

\end{document}